\documentclass[letterpaper, 10 pt, conference]{ieeeconf}  

\IEEEoverridecommandlockouts                              

\usepackage{graphicx} 
\usepackage{siunitx}
\usepackage[caption=false, font=footnotesize]{subfig}
\usepackage{nicefrac}
\usepackage{booktabs}
\usepackage{amsmath} 
\usepackage{mathrsfs}
\usepackage{xcolor}
\usepackage{gensymb}
\usepackage{tikz}
\usepackage{url}
\title{\LARGE \bf
Minimalist Visual Inertial Odometry
}

\newcommand\submittedtext{%
  \footnotesize This work has been submitted to the IEEE for possible publication. Copyright may be transferred without notice, after which this version may no longer be accessible.}

\newcommand\submittednotice{%
\begin{tikzpicture}[remember picture,overlay]
\node[anchor=south,yshift=10pt] at (current page.south) {\fbox{\parbox{\dimexpr0.65\textwidth-\fboxsep-\fboxrule\relax}{\submittedtext}}};
\end{tikzpicture}%
}

\author{
Francesco Pasti, Jeremy Klotz, Nicola Bellotto, and Shree K. Nayar
\thanks{F. Pasti and N. Bellotto are with the Department of Information Engineering, University of Padua, Padua, Italy. F. Pasti, J. Klotz, and S. K. Nayar are with the Computer Science Department, Columbia University, New York, NY, USA. Corresponding author: {\tt\small francesco.pasti@dei.unipd.it}}%
}

\begin{document}

\maketitle
\submittednotice
\thispagestyle{empty}
\pagestyle{empty}

\begin{abstract}
Visual-Inertial Odometry~(VIO), which is critical to mobile robot navigation, uses cameras with a large number of pixels. Capturing and processing camera images requires significant resources. 
This work presents a minimalist approach to planar odometry, showing that just four visual sensors (pixels) and an IMU provide robust motion estimation for differential-drive robots.
Our key insight is that four downward-facing photodiodes that sense the world through optical Gabor masks produce signals that encode speed. 
Based on this, we jointly optimize the mask parameters alongside a Temporal Convolutional Network~(TCN) using a physically-grounded simulator.
The resulting model decodes speed from the four photodiode measurements.
Pairing these estimates with an IMU's angular speed yields a continuous planar trajectory. 
We validate our approach with a prototype sensor mounted on a differential drive robot. 
Across diverse indoor and outdoor terrains, our system closely tracks the reference trajectories without any real-world fine-tuning. 
Our work shows that minimalist sensing enables efficient and accurate planar odometry.
\end{abstract}

\section{Introduction}\label{sec:intro}
Autonomous mobile robots often rely on Visual-Inertial Odometry~(VIO) for robust navigation~\cite{carlone2025slam}. VIO fuses rich visual cues with measurements from an Inertial Measurement Unit~(IMU) to compute the trajectory of the robot. These visual cues are extracted from camera images with a large number of pixels. Since the power consumption to sense and process images is roughly linear in the number of pixels, traditional VIO using high resolution image sensors can be unsuitable for resource-constrained platforms~\cite{likamwa2013energy,neuman2022tiny}. 

\begin{figure}
        \centering
        \includegraphics[width=\linewidth]{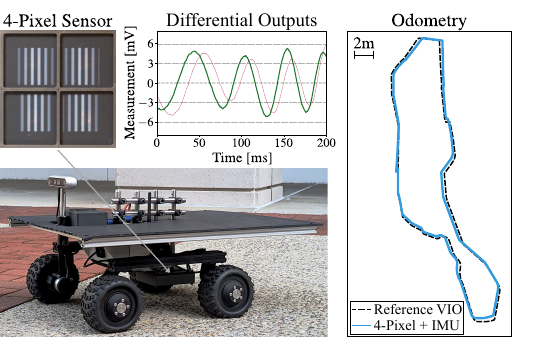} 
        \vspace{-0.4in}
        \caption{\textbf{The minimalist odometry system utilizes a custom sensor consisting of four masked photodiodes}. The masks act as analog spatial filters that isolate specific spatial frequencies from the ground texture. As the differential drive robot moves, this optical filtering generates continuous temporal signals. We decode these signals to regress the robot's instantaneous speed. Fusing this speed with the IMU's gyroscope yaw rate yields the full planar trajectory.}
        \label{fig:teaser}
        \vspace{-0.2in}
\end{figure}

Our work draws inspiration from minimalist vision~\cite{pooj2018minimalist,klotz2024minimalist}, which explores the lower bound of visual information needed to solve a vision task.
We show that for differential drive robots, robust planar odometry can be achieved using just four ground-facing pixels, where each pixel is a photodiode with an optical mask~(Fig.~\ref{fig:teaser}). We show that when the 
masks represent Gabor functions, i.e., sinusoidal waves modulated by a Gaussian envelope~\cite{gabor1946theory}, the masks isolate a specific spatial frequency from the ground texture. As the robot moves, the sensor generates a temporal signal whose dominant frequency directly encodes the robot's linear speed.

In any real-world setting, odometry has to contend with non-linear motion dynamics, unknown and varying ground textures, and hardware noise. Therefore, decoding the signals produced by our four Gabor pixels is a non-trivial problem. To achieve this, we develop an end-to-end differentiable framework that jointly optimizes the Gabor mask parameters alongside a Temporal Convolutional Network~(TCN) decoder~\cite{lea2017temporal}.
This system is trained using data produced by a physically grounded simulator that uses a diverse set of real-world textures and motion profiles. The end result is a robust mapping of our pixel measurements to linear speed. This speed estimate is fused with the yaw rate from an IMU's gyroscope to obtain the full planar trajectory of the robot.

We developed a hardware prototype of our sensor that we mounted on a differential drive robot.
To validate our approach, we drove the robot for 87~minutes over 920~meters of diverse indoor and outdoor terrains.
Even though our minimalist system has been optimized purely on simulated data, it achieves robust real-world odometry that closely tracks the reference trajectories computed via standard high-resolution VIO.
Our method achieves a mean absolute trajectory error~(ATE) of 0.34~meters and an average endpoint drift of 0.60\%.
In contrast, standard wheel encoder and IMU fusion yields a 0.74~meters ATE and a 1.55\% drift.
This performance is achieved despite the significant reduction in sensing resources of our 4-pixel sensor compared to high-resolution VIO, demonstrating that minimalist vision can be a viable solution for resource-constrained robot odometry.

\section{Related Works}\label{sec:related_works}

\subsection{Odometry in Robotics}
Robust mobile robot localization relies on the fusion of proprioceptive sensors like wheel encoders and IMUs with exteroceptive sensors like cameras and LiDARs~\cite{carlone2025slam}.
Wheel encoders are a reliable short-term baseline but drift under wheel slippage. IMUs offer high-frequency and low-power proprioception, but double integrating their noisy acceleration causes a large position drift over time.
While data-driven methods can constrain this drift for pedestrian dead-reckoning~\cite{liu2020tlio}, they fail to generalize to the smooth, non-oscillatory kinematics of wheeled robots~\cite{mohamed2019survey}.

To mitigate these errors, Visual-Inertial Odometry~(VIO) is the standard solution, fusing exteroceptive visual cues with IMU data to deal with rapid maneuvers and unreliable visual information~\cite{zhao2025resilient}.
However, conventional VIO requires processing video streams of thousands or millions of pixels per frame. The resulting computational and energy requirements are prohibitive for resource-constrained platforms operating autonomously over extended periods~\cite{neuman2022tiny}.
Learning-based VIO methods further improve accuracy through dense flow prediction and multi-source fusion~\cite{liu2025enhancing, zhang2026multi, liu2026fso}.
In contrast, our framework uses only 4 pixels and an IMU, greatly reducing sensing resources while maintaining robust odometry performance.

\subsection{Efficient Optical Flow Estimation}
Another approach to motion estimation uses optical flow sensors adapted for robotics~\cite{honegger2013open}.
These systems use small pixel arrays ($18\times18$ to $30\times30$) and dedicated signal processors to compute spatial cross-correlation between high-framerate images.
While optimized for low latency, digitizing and processing 2D images thousands of times per second limits their efficiency.
These downward-facing sensors are also highly sensitive to their standoff distance from the ground, which should be fixed with millimeter accuracy. This limits their robustness on uneven terrain, where chassis vibrations make the standoff distance vary continuously.

Similarly, event cameras offer a low-power alternative for motion estimation~\cite{mueggler2018continuous}. Their sparse, asynchronous events encode per-pixel brightness changes, drastically reducing data rates while maintaining high temporal resolution.
Nevertheless, extracting motion from these event streams requires dense spatial sampling of the scene.

Rather than computing optical flow digitally, our approach draws on a model of biological motion perception, which shows that spatio-temporal frequency filtering of light measurements directly yields motion estimation~\cite{land1992evolution,adelson1985spatiotemporal}.
We implement this principle in optics.
As the robot moves, our sensor optically convolves the scene texture with Gabor filters to produce a low-dimensional temporal signal that encodes motion in the frequency domain. We show that our system achieves comparable accuracy to optical flow based methods that use orders of magnitude more pixels.


\subsection{Minimalist Vision}
Minimalist vision aims to sense the smallest number of task-relevant visual measurements without capturing a full image. For instance, the minimalist camera in~\cite{pooj2018minimalist} uses handcrafted optical masks over photodetectors to process scene information in the optical domain.
Freeform pixels~\cite{klotz2024minimalist} extend this idea, modeling masked photodetectors as linear projectors of the scene. Such physical masks can be included as the initial layer of a neural network. Training it for a given task yields the learned freeform pixels and corresponding inference network.

Our approach builds on this framework with a critical distinction. Since motion estimation is inherently based on spatio-temporal frequency analysis~\cite{adelson1985spatiotemporal}, Gabor filters are ideally suited to isolate the motion-related frequency components. Rather than designing freeform pixels for speed estimation, which is not well constrained, we constrain the masks to be Gabor functions with learnable parameters. These are optimized for the complex real-world conditions (texture, lighting, mechanical) faced by the sensor. The outputs of our Gabor pixels feed a TCN decoder, and the Gabor parameters and the TCN are jointly trained to obtain a robust speed estimator. 



\section{The Minimalist Odometry System}

\subsection{Theoretical Intuition}\label{sec:theory}
\begin{figure}
    \centering
    \includegraphics[width=\linewidth]{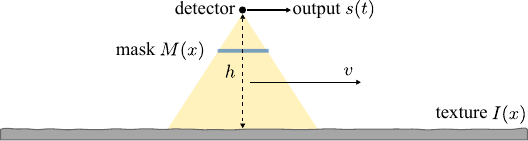}
    \vspace{-0.21in}
    \caption{\textbf{Theoretical intuition}. A detector integrates the light from a surface texture $I(x)$ passing through an optical mask $M(x)$. As the sensor moves at a constant speed $v$, it performs a continuous spatial cross-correlation in the optical domain. This physical process maps the scene's spatial frequencies into a temporal signal $s(t)$.}\label{fig:signal_theory}
    \vspace{-0.18in}
\end{figure}

Consider the simplified scenario illustrated in Fig.~\ref{fig:signal_theory}. A sensor, consisting of a single detector and an optical mask, is pointed down at a surface and moves over it along its longitudinal $x$-axis with a constant speed $v$. While the surface texture is two-dimensional, let the optical mask have translational symmetry along the lateral $y$-axis.
This reduces their optical interaction to a single dimension, allowing us to model the system strictly as one-dimensional along the direction of motion.

Let the texture of the surface along the direction of motion be the function $I(x)$ and the transmittance function of the mask, when projected onto the surface, be $M(x)$.
If the sensor is positioned at an arbitrary displacement $x$, the detector integrates the light passing through the mask at that location. Therefore, the sensor output as a function of position is simply the spatial cross-correlation of the texture and the mask: $S_x(x) = (I \star M)(x)$.

If the sensor moves at a constant speed $v$, the detector's displacement relative to the surface is $x=vt$.
Therefore, the continuous temporal output $s(t)$ of the sensor is a scaled version of the spatial cross-correlation $S_x(x)$, such that $s(t) = S_x(vt)$.
This directly links the temporal and spatial domains via $x=vt$.

To understand how this temporal signal $s(t)$ encodes speed, we analyze it in the frequency domain.
Let $f$ denote temporal frequency and $\mathcal{F}$ denote the Fourier transform.
Then, the transform of the sensor's temporal output is $\hat{S}(f) =\mathcal{F}\{S_x(vt)\}$.
By applying the time-scaling property of the Fourier transform, we get:
\begin{equation}\label{eq:one}
 \hat{S}(f) = \frac{1}{|v|}\hat{S}_x\left(\frac{f}{v}\right).
\end{equation}
Now let $\xi$ denote frequency in the spatial domain.
Using the cross-correlation theorem, the Fourier transform of $S_x(x)$ can be expressed as $\hat{S}_x(\xi)=\hat{I}(\xi)\hat{M}^*(\xi)$, where
$*$ denotes the complex conjugate, and $\hat{I}$ and $\hat{M}$ are the transforms of the texture and the mask, respectively.
By substituting $\hat{S}_x(\xi)$ in Eq.~\ref{eq:one}, we get:
\begin{equation}\label{eq:two}
 \hat{S}(f) = \frac{1}{|v|}\hat{I}\left(\frac{f}{v}\right)\hat{M}^*\left(\frac{f}{v}\right).
\end{equation}
The above equation represents the core working principle of our sensor. 
The temporal frequencies ($f$) produced by the moving sensor are the spatial frequencies ($\xi$) of the texture, filtered by the mask, and scaled by the sensor speed $v$, such that $f=\xi v$.

Our challenge is to estimate the sensor speed $v$ robustly, irrespective of the texture $I(x)$ of the surface. 
Consider an infinite cosine mask $M(x) = \cos(2 \pi \xi_0 x)$, where $\xi_0$ is the known frequency of the cosine.
In the frequency domain, its spectrum $\hat{M}(\xi)$ consists of two symmetric impulses at $\pm\xi_0$. Although the spectrum of the texture $\hat{I}(\xi)$ is unknown, we can assume it to be broadband. Then, our ideal mask has the effect of passing through to the sensor a single spatial frequency $\xi_0$ from the texture. 
In Eq.~\ref{eq:two}, this makes the spectrum of the sensor output $\hat{S}(f)$ collapse to symmetric impulses at $\pm f_0$, where $f_0=\xi_0 v$. Therefore, if we can detect $f_0$ from $s(t)$, the speed can be estimated as  $v=\nicefrac{f_0}{\xi_0}$.

While an infinite cosine mask provides a theoretical baseline, any physical mask must have a finite aperture. In the frequency domain, a cosine mask with an aperture is no longer a pair of impulses at $\pm\xi_0$, but rather a pair of broader functions, which makes the estimation of speed harder.

To mitigate this broadening of the impulses, we make the mask a Gabor function~\cite{gabor1946theory}:
\begin{equation}\label{eq:three}
     G_{cos}(x) = \alpha \exp\left(-\frac{x^2}{2\sigma^2}\right) \cos(2 \pi \xi_0 x) ,
\end{equation}
where $\alpha$ is an amplitude scaling factor and $\sigma^2$ is the variance of a Gaussian envelope.
By modulating the mask with a Gaussian envelope, we smoothly taper the cosine to zero, avoiding its abrupt spatial truncation.
So long as the Gaussian envelope is broad, the spectrum of the Gabor mask has narrow peaks at \(\pm\xi_0\).
Similar to the ideal cosine, the Gabor mask restricts the spectrum of $\hat{S}(f)$ in Eq.~\ref{eq:two} to symmetric peaks at $\pm f_0$, where $f_0=\xi_0 v$
The resulting temporal sensor output $s(t)$ can be approximated as an amplitude-modulated cosine whose frequency remains $f_0 = \xi_0 v$.
This allows us to recover the speed $v$ by finding the fundamental frequency $f_0$ of the sensor's output $s(t)$.

This brings us to a limitation of the above method.
If the sensor reverses its direction of motion, the speed $v$ becomes $-v$.
This theoretically yields a negative frequency $f_0 = \xi_0 (-v)$ in the sensor output $s(t)$.
However, since $s(t)$ is a real-valued signal, it still contains symmetric peaks at $\pm f_0$. Therefore, this method would only determine the magnitude of the speed (the fundamental frequency $f_0$), not the direction of motion (the sign of $v$).


\subsection{Directional Ambiguity and Positive Masks}\label{subsec:theory-quad}

To resolve the above directional ambiguity, we introduce a second masked sensor, co-located and in quadrature with the first.
Let its mask be a sine Gabor function, $G_{sin}(x)$, which is $G_{cos}(x)$ in~Eq.~\ref{eq:three} phase shifted by $90\degree$, corresponding to a translation in space.
In the frequency domain, the sine Gabor also has peaks at $\pm\xi_0$.
However, the spatial shift between the two masks naturally produces a $90\degree$ temporal phase shift between the outputs of the two respective sensors.

Let the outputs of the two sensors be \(s_{cos}(t)\) and \(s_{sin}(t)\).
Since we know that both outputs are narrowband with peaks at $f_0 = \xi_0 v$, we can approximate them as: 
\begin{equation}\label{eq:signals}
\begin{aligned}
s_{cos}(t) &\approx A(t) \cos(2\pi\xi_0vt + \theta(t)) , \\
s_{sin}(t) &\approx A(t) \sin(2\pi\xi_0vt + \theta(t)) .
\end{aligned}
\end{equation}
Here, $A(t)$ and $\theta(t)$ represent the time-varying amplitude and spatial phase of the texture $I(x)$ at $\xi_0$.
Crucially, the fundamental frequencies of the two signals are the same, $f_0=\xi_0 v$.

Intuitively, the sign (direction) of the speed $v$ is encoded in the relative phase between \(s_{cos}(t)\) and \(s_{sin}(t)\). Note that this relative phase is either $+90\degree$ or $-90\degree$, one corresponding to forward motion and the other to backward motion of the sensors, while, as before, $f_0$ reveals the speed magnitude. These results generalize to dynamic motion as well. Even when the speed varies with time, the frequency and relative phase represent the speed and direction at that time. 

There is an additional constraint we need to consider when building our sensor. While Gabor filters have positive and negative values, the optical transmittance of a mask can only be positive. We therefore decompose each Gabor mask function into a pair of strictly non-negative masks:
\begin{equation} 
    M^+ = \max(G(x), 0), \quad M^- = \max(-G(x), 0) \,.
\end{equation}
The difference between the outputs of two sensors with the above masks is equivalent to the output of a single sensor with an idealized Gabor mask.
Therefore, in order to obtain the two quadrature sensor outputs, we use four sensors with the masks $M_{cos}^+$, $M_{cos}^-$, $M_{sin}^+$, and $M_{sin}^-$. This is the basic design of our minimalist sensor. 

\subsection{Height Dependency}\label{subsec:height}

The above derivation of minimalist sensing for speed estimation is valid when the four masked detectors are co-located. In practice, however, they must be spatially offset, as shown in Fig.~\ref{fig:hardware}. Given the nominal height $h_{nom}$ of the masks from the ground plane, the masks are positioned with respect to their detectors such that their projections onto the ground plane are perfectly aligned. In other words, the projected spatial frequencies of the four masks are the same ($\xi_0$), and the temporal phase shift between the quadrature signals is exactly $90\degree$. This makes the system equivalent to the co-located one in Sec.~\ref{subsec:theory-quad} for the height $h_{nom}$.

In any real setting, however, due to the unevenness of the ground and the robot's vibrations, the height of the sensor will vary ($h=h_{nom} + \Delta h$). This changes the scales of the projections of the masks onto the ground plane, and hence their effective spatial frequency ($\xi_0 + \Delta \xi_0$). It also introduces parallax between the offset detectors, which no longer observe the same patch on the ground plane. While this corrupts the sensor outputs, it also provides a latent cue. The spatial displacement between the projected masks changes as a function of height, so the phase difference between $s_{cos}$ and $s_{sin}$ encodes information regarding deviations from $h_{nom}$.

These height dependent phase shifts are confounded by the random undulations of the ground plane, and their effect on the speed estimate is hard to model analytically. We are able to exploit them nonetheless using our learning framework~(Sec.~\ref{sec:learning}), by training the model on simulated sequences with realistic textures and height variations.

\subsection{Planar Kinematics and IMU Integration}\label{subsec:theory-planar}
Our goal is to recover the planar odometry of a differential-drive robot.
We have shown that our minimalist sensor encodes the signed speed $v$ along its direction of motion. We are therefore able to measure the robot's speed $v_x$ in the ``forward’’ direction. We mount our sensor with its center on the longitudinal axis of the robot and its mask stripes perpendicular to the direction of motion (see Fig.~\ref{fig:hardware}(a)).
Note that there are a few conditions for which the speed estimation from the detector outputs can be challenging. For instance, when the robot undergoes simultaneous rotation and translation, the speed of the ground plane is non-uniform within the sensor's field of view. Also, if the robot experiences lateral slippage, the detector outputs are impacted by the variation in ground texture due to the slip. To deal with these effects, our model for forward speed estimation is trained using a simulator that includes these motion dynamics.

Full planar odometry also requires the rotational component, i.e., the yaw rate $\omega_z$. To this end, we integrate an IMU whose gyroscope independently measures the yaw rate, $\omega_z$. 
By fusing our optically estimated forward speed $v_x$ with the IMU's yaw rate $\omega_z$, we obtain the full planar odometry through the open-loop integration of the unicycle model in which the robot follows a circular arc of radius $\nicefrac{v_x}{\omega_z}$~\cite{siegwart2011introduction}.

\subsection{Sensor Prototype}\label{subsec:prototype}
\begin{figure}
    \centering
    \vspace{-0.1in}
    \captionsetup[subfloat]{captionskip=0pt}
    \subfloat[Masks\label{subfig:masks}]{\includegraphics{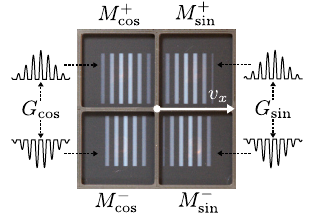}}
    \subfloat[Photodiodes\label{subfig:pcb}]{\includegraphics{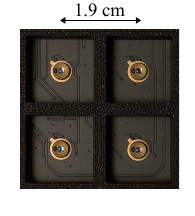}}
    \vspace{-0.05in}
    \caption{\textbf{Hardware implementation of our 4-pixel speed sensor.} (a) Since light transmission is strictly positive, the Gabor filters ($G_{cos}$ and $G_{sin}$) are split into their positive and negative components to obtain the masks $M^+$ and $M^-$. The resulting four masks are printed on film. (b) They are placed in front of a $2\times2$ grid of photodiodes. The distance between adjacent photodiodes is $1.9~\text{cm}$.}\label{fig:hardware}
    \vspace{-0.18in}
\end{figure}

We designed our minimalist sensor based on the custom hardware architecture developed for freeform pixels~\cite{klotz2024minimalist}.
The four masks $M_{cos}^+$, $M_{cos}^-$, $M_{sin}^+$, and $M_{sin}^-$, each $16\times16~\text{mm}^2$ in size, are printed on transparent film~(Fig.~\ref{fig:hardware}(a)).
The masks are placed $11.4~\text{mm}$ above four Hamamatsu S9119-01 photodiodes arranged in a $2\times2$ grid~(Fig.~\ref{fig:hardware}(b)). This results in a $\ang{70}$ field of view for each detector. We have positioned the masks to ensure that the four detectors observe the same area on the ground plane at a nominal height $h_{nom}=6~\text{cm}$.

In our experiments, we used an external data acquisition~(DAQ) system to digitize the four analog signals produced by the above sensor. Our sensor produces four analog detector outputs consuming just $2.5~\text{mW}$. In comparison, the image sensor in a conventional camera consumes hundreds of milliwatts~\cite{likamwa2013energy}. This translates to a reduction of two orders of magnitude in the power consumed by the sensor.

\section{Learning Framework for Speed Estimation}\label{sec:learning}
\begin{figure*}
    \centering
    \includegraphics[width=\linewidth]{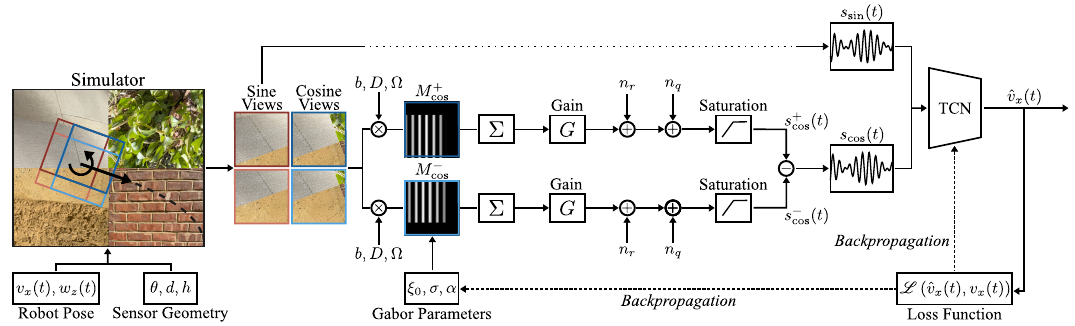}
    \vspace{-0.25in}
    \caption{\textbf{Overview of our learning framework.} The simulator generates the four detector views $I_{k}(x,y,t)$ of our sensor during kinematic motion. The effects of the finite area of the detector ($b$), the directional response of the detector ($D$) and the foreshortening effect ($\Omega$) are applied to the detector views before they are modulated by the masks $M_{k}(x,y,t)$. These modulated views are integrated ($\Sigma$) to get four signals to which a gain ($G$), read noise ($n_r$), quantization noise ($n_q$) (of the DAQ) and saturation are applied. From these four detector outputs, $s_{\cos}(t)$ and $s_{\sin}(t)$ are computed and fed into a TCN decoder. The TCN takes a temporal window of the input signals to predict the linear speed $\hat{v}_x(t)$ of the sensor. Since the pipeline is fully differentiable, the loss $\mathscr{L}$ is backpropagated to optimize not just the TCN parameters but also the Gabor parameters~($\xi_0, \sigma, \alpha$).}
    \label{fig:simulator}
    \vspace{-0.18in}
\end{figure*}

Finding a robust mapping from our minimalist sensor signals to speed  is a non-trivial problem. The theoretical model described in Sec.~\ref{sec:theory} must cope with various factors including uneven ground, complex robot motions, and noise in the sensor outputs. Furthermore, although we have established our masks will be Gabor functions, we have not yet determined what their parameters~($\xi_0, \sigma, \alpha$) should be. This is a challenging problem as it depends on the properties of the wide range of textures the sensors will encounter. 

We address the above problems using a learning-based approach, where the Gabor parameters and the parameters of a network for speed estimation are jointly learned with simulated sensor data. To this end, we built a physically based simulator~(Fig.~\ref{fig:simulator}) that generates sensor outputs for a wide range of robot motions and ground plane textures. 

\subsection{Minimalist Odometry Sensor Simulation}\label{subsec:diff_sim}

Fig.~\ref{fig:simulator} shows the complete pipeline of our learning framework. We generate realistic training data for our speed estimator using a variety of high-resolution surface textures and dynamic kinematic trajectories. To ensure our speed estimator can cope with diverse (indoor and outdoor) environments, our simulator uses 
real-world textures from the Matador dataset~\cite{beveridge2025hierarchical}, comprising $\sim 7200$ high-quality images across 57 material categories.
For realistic movements, we simulate our minimalist sensor across over $80~\text{km}$ and $12~\text{h}$ from the TartanGround~\cite{patel2025tartanground} dataset, encompassing linear speeds up to $ 5~\nicefrac{\text{m}}{\text{s}}$ and angular speeds up to $1~\nicefrac{\text{rad}}{\text{s}}$.
To avoid aliasing, we upsample the $100~\text{Hz}$ pose trajectories in \cite{patel2025tartanground} to $1000~\text{Hz}$.

At each timestep $t$, the robot pose defines the spatial coordinates of the four detectors, indexed by $k \in \{\cos^+,\cos^-,\sin^+, \sin^-\}$. The sensor geometry~(Sec.~\ref{subsec:prototype}) is parametrized by its inter-detector distance $d$ and field of view $\theta$.
We randomly perturb the nominal sensor height ($h=h_{nom}+\Delta h$) to simulate the scale error~(Sec.~\ref{subsec:height}).

For each detector, we use resampling with bilinear interpolation to construct a $128\times128$ pixel image from the high-resolution texture in its field of view. The resulting images $I_k(x,y,t)$, corresponding to the four detectors, are converted to grayscale and their brightness values are  normalized to~$[0,1]$. We also render the aligned Gabor masks $M_{k}(x,y)$ as $128\times128$ images, parametrized by the spatial frequency $\xi_0$, sigma $\sigma$, and amplitude $\alpha$ of the Gabor function. Since our physical masks are printed on transparency film, we constrain their transmittance to lie within $[0,1]$. 

Using the image patches $I_k(x,y,t)$, we simulate the optical pipeline of our sensor. Following the physical sensor model described in~\cite{klotz2024minimalist}, we first blur $I_{k}$ with a kernel $b(x,y)$, which represents the finite size of the detector. Then, we multiply the resulting image with the directional response of the detector $D(x,y)$ and $\Omega(x,y)$, which accounts for the foreshortening effect. The resulting image is multiplied with the Gabor mask $M_k(x,y)$ and then integrated to obtain the final detector output. This entire process of simulating a detector output can be mathematically written as:
\begin{equation}
    s_{k}(t)=\sum_{\mathsf{u}}\Big[I_{k}(\mathsf{u},t)*b(\mathsf{u})\Big]D(\mathsf{u})\,\Omega(\mathsf{u})\,M_{k}(\mathsf{u}) \,,
\end{equation}
where $s_{k}$ is the detector output and $\mathsf{u} = (x,y)$.

Finally, we need to convert our sensor output $s_{k}(t)$ to a voltage. As in~\cite{klotz2024minimalist}, we scale $s_{k}(t)$ by a calibrated hardware gain $G=1.22\times10^{-4}$ and inject both Gaussian read noise~($\sigma_r=175~\mu\text{V}$) and uniform quantization noise. 
The result is then clipped at $3.2~\text{V}$ to model the finite dynamic range of the detectors.
To construct the final inputs for our speed decoder, we compute $s_{cos}^+(t) - s_{cos}^-(t)$ and $s_{sin}^+(t) - s_{sin}^-(t)$, obtaining $s_{cos}(t)$ and $s_{sin}(t)$, respectively.

\subsection{Temporal Speed Decoder and End-to-End Optimization}\label{subsec:tcn}

To estimate the sensor's linear speed, we process a \mbox{1-second} observation window with 1000 temporal samples of our signals $s_{cos}(t)$ and $s_{sin}(t)$~(i.e. $2\times1000$ input matrix), using a TCN~\cite{lea2017temporal} with 184K parameters. The TCN consists of 9 residual blocks with kernel size 3 and dilation doubling from 1 to 256, with channel widths of 16, 32 and then 64, group normalization and dropout, yielding a receptive field that covers the 1-second input window. The TCN extracts multi-scale temporal features via stacked dilated convolutions and an attention pooling layer.

Since the amplitudes~($A(t)$ in Eq.~\ref{eq:signals}) of the signals $s_{cos}(t)$ and $s_{sin}(t)$ depend on the spectrum of the ground plane texture, the signal-to-noise ratio~(SNR) of the signals can vary within our 1-second observation window. The 
attention pooling layer of the TCN mitigates this problem by allowing the network to suppress noisy portions of the window.
The pooled features feed into a residual fully-connected head, which outputs a linear speed prediction $\hat{v}_x(t)$ and a log-variance $\log(\hat{\sigma}^2)$ that quantifies prediction uncertainty.

We train the network with a Gaussian negative log-likelihood loss, which optimizes both the speed prediction and its uncertainty.
To make training stable under high accelerations, we define the target speed $v_x(t)$ as the average sensor speed over the last $0.1~\text{s}$ of the observation window.

Since our simulator~(Sec.~\ref{subsec:diff_sim}) is fully differentiable, the loss gradients propagate from the TCN back to the simulated sensor. This lets us jointly optimize the parameters of the TCN and of our Gabor masks ($\xi_0, \sigma$, and $\alpha$). In other words, we are able to jointly design our sensor (hardware) and speed estimator (software) to achieve high speed accuracy.

Because the TCN dominates the computational cost of our system, we assess its real-time feasibility by deploying it on embedded hardware. On a Raspberry Pi 3B, the TCN requires $176~\text{MMAC}$ per inference within $2.1~\text{MB}$ of peak memory, and a full inference takes $30.3~\text{ms}$, corresponding to an update rate of $33~\text{Hz}$. This demonstrates that our system can run in real time on an embedded platform.

\section{Experimental Evaluation}\label{sec:experiments}

\subsection{Mask Design and Component Ablation}\label{subsec:mask_ablation}

We evaluated in simulation the impact of the mask parameterization on speed estimation accuracy. We refer to the Gabor masks jointly optimized with the TCN as \textit{Learned Gabor}. We compared them with \textit{Fixed Gabor} masks, where the Gabor parameters were empirically selected to be $\xi_0=6$, $\sigma = 1$, and $\alpha = 1$. We also compared them with \textit{Freeform Pixels}~\cite{klotz2024minimalist}, whose masks have arbitrary transmittance functions that are jointly optimized with the TCN.

In these experiments, we used an inter-detector distance of $d=1.9~\text{cm}$ and detector field of view of $\theta=\ang{70}$ (see Sec.~\ref{subsec:diff_sim}). We set the nominal height to $h_{nom}=20~\text{cm}$ to prevent aliasing at the high speeds ($\pm 5\nicefrac{\text{m}}{\text{s}}$) present in the TartanGround dataset.
We split the Matador textures and TartanGround trajectories into train~($70\%$), validation~($10\%$), and test~($20\%$) sets.
The three mask designs described above and the TCN were trained for $100$ epochs using Adam optimizer with learning rate of $10^{-4}$ and batch size of $32$.

Table~\ref{tab:ablation} summarizes these configurations in terms of the root-mean-square error~(RMSE) and mean absolute error~(MAE) in the estimated speed. \textit{Freeform Pixels} perform poorly. Without the Gabor prior, the masks converge to blurred patterns, forcing the TCN to rely on the temporal lag between the displaced detectors' signals to regress speed.
Gabor masks significantly improve performance.
In particular, \textit{Learned Gabor} masks yield a $29\%$ lower RMSE and $35\%$ lower MAE than the \textit{Fixed Gabor} masks.

Next, we conduct ablations to illustrate the importance of the quadrature masks, the Gaussian envelope, and the TCN, as summarized in the bottom three rows of Table~\ref{tab:ablation}. \textit{No sine masks} refers to a single Gabor mask pair without the additional quadrature pair. As expected, this configuration achieves poor speed estimation since the measured signal only encodes the magnitude of the speed, not the direction. Training the Gabor masks without the Gaussian envelope in Eq.~\ref{eq:three} also yields poor speed estimation due to the spectral leakage in the measured signals caused by the masks' finite apertures. Finally, we replaced the TCN with an analytical decoder that estimates speed by directly fitting the phase and frequency in Eq.~\ref{eq:signals} to the measured Gabor signals. This approach is not robust to non-uniform textures or sensor noise, and it fails to accurately measure speed in the presence of acceleration or rotation. Since the TCN is trained using simulated data that models all of these effects, it achieves a lower error in the estimated speed.


\begin{table}[t]
    \centering
    \caption{Ablation on Mask Designs and System Components}\label{tab:ablation}
    \vspace{-0.1in}
    \begin{tabular}{@{}l c c@{}}
        \toprule
        \textbf{Configuration} & \textbf{RMSE ($\nicefrac{\text{m}}{\text{s}}$)} & \textbf{MAE ($\nicefrac{\text{m}}{\text{s}}$)} \\
        \midrule
        Learned Gabors (ours)                & \textbf{0.054} & \textbf{0.034} \\
        \midrule
        Fixed Gabors                         & 0.076 & 0.052 \\
        Freeform                             & 0.147 & 0.101 \\
        \midrule
        No sine masks       & 1.988 & 1.651 \\
        No Gaussian envelope & 1.910 & 1.500 \\
        Analytical decoder   & 1.118 & 0.743 \\
        \bottomrule
    \end{tabular}
    \vspace{-0.2in}
\end{table}


\subsection{Robustness to Height Variations}\label{subsec:height_exp}
As discussed in Sec. \ref{subsec:height}, height deviations alter the effective spatial frequency of the Gabor~($\xi_0~+~\Delta~\xi_0$), impacting the frequency-to-speed mapping. However, height variations also change the phase difference between the detector outputs $s_{cos}$ and $s_{sin}$, which is a useful cue. 

We evaluated via simulations the TCN's ability to exploit this phase difference to achieve robustness to height variations~($h_{nom}+\Delta h$). We used the same sensor configuration parameters as in Sec.~\ref{subsec:mask_ablation}, but we trained and tested the \textit{Learned Gabor} masks and the TCN with the random height variations uniformly sampled from within the ranges of $\pm 10\%$, $\pm 25\%$ and  $\pm 50\%$, relative to $h_{nom}= 20~\text{cm}$.

As shown in Table~\ref{tab:sim_height_results}, the performances of the model trained at a fixed height $h_{nom}$ degrade significantly under height variations. In contrast, training with the uniform height randomization makes speed estimation resilient to them. 
A $\pm 25\%$ uniform height randomization during training yields robust performance for the same range of height variation during testing. In fact, it slightly improves the accuracy at the nominal height, yielding a speed RMSE of $0.048~\nicefrac{\text{m}}{\text{s}}$.
This shows that exposing the model to these variations during training results in a speed estimator that implicitly learns the relationship between height, phase, frequency and speed.

However, when trained and tested on substantial height variations~($\pm 50\%$), extreme parallax between the views seen by the four detectors makes the phase cue unreliable and the overall performance drops.
In our real-world experiments, we found that training with the $\pm 25\%$ uniform height randomization results in a system that is resilient to uneven terrains and vibrations encountered in both indoor and outdoor settings.

\begin{table}[t]
    \centering
    \caption{Speed RMSE~($\nicefrac{m}{s}$) as height deviates from $h_{nom}=20~\text{cm}$}
    \vspace{-0.1in}
    \label{tab:sim_height_results}
    \begin{tabular}{@{}l c c c c @{}}
        \toprule
        & \multicolumn{4}{c}{\textbf{Test Height Range}} \\
        \cmidrule(lr){2-5}
        \textbf{Train Height Range} & $h_{nom}$ & $\pm 10\%$ & $\pm 25\%$ & $\pm 50\%$ \\
        \midrule
        $h_{nom}$          & 0.054 & 0.070 & 0.266 & 1.542 \\
        $h_{nom} \pm 10\%$ & 0.050 & 0.054 & 0.063 & 1.154 \\
        $h_{nom} \pm 25\%$ & \textbf{0.048} & 0.050 & 0.056 & 1.259
        \\
        $h_{nom} \pm 50\%$ & 0.071 & 0.071 & 0.075 & 0.088 \\
        \bottomrule
    \end{tabular}
    \vspace{-0.2in}
\end{table}

\subsection{Robot with Minimalist Odometry Sensor}\label{subsec:real-robot}
\begin{figure}
        \centering
        \includegraphics[width=\linewidth]{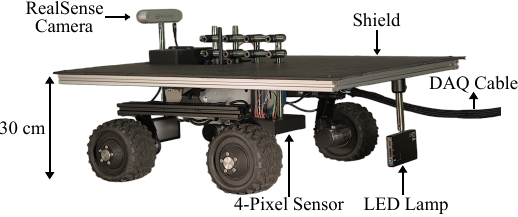}
        \vspace{-0.3in}
        \caption{\textbf{Differential drive robot with minimalist odometry sensor}. Our 4-pixel sensor is mounted downward-facing at a nominal height of $h_{nom}=6~\text{cm}$ on a differential-drive robot. A shield is used to suppress strong specular reflections from the ground plane, and an LED lamp provides illumination in poorly lit indoor environments. An Intel RealSense D455 camera is used for the reference VIO trajectories. The IMU of this camera provides the angular speed which is fused with our sensor's linear speed estimates to compute planar odometry.}
        \label{fig:robot}
        \vspace{-0.25in}
\end{figure}

We evaluated our minimalist odometry sensor prototype~(Fig.~\ref{fig:hardware}) on the LeoRover 
differential-drive robot shown in Fig.~\ref{fig:robot}. We mounted the sensor at a height 
$h_{nom}$ of $6~\text{cm}$, 
centered on the robot's longitudinal axis to measure its linear speed $v_x$.
A shield avoids strong specular reflections from light sources in the environment towards the sensor.
To cope with low light indoors, we used a low-powered LED source suspended from the chassis of the robot. This source is not used outdoors or in well-lit indoor environments.

We jointly trained the Gabor masks and the TCN with a nominal sensor height $h_{nom}$ of $6~\text{cm}$ and a uniform height randomization of $\pm 25\%$. We scaled the speed in the TartanGround trajectories to match the maximum speed of the robot~($0.4~\nicefrac{\text{m}}{\text{s}}$). The resulting model has a speed RMSE of $0.004~\nicefrac{\text{m}}{\text{s}}$ and MAE of $0.003~\nicefrac{\text{m}}{\text{s}}.$
The trained masks were printed on transparency film and installed on the prototype without further modifications.

An external DAQ samples the four detectors signals at $41.6~\text{kHz}$. Then, we process the resulting signals with a $60~\text{Hz}$ notch filter to remove AC lighting fluctuations and a $450~\text{Hz}$ low-pass filter to denoise the signals before downsampling them to $1~\text{kHz}$, which is the temporal resolution the TCN is designed for.

\subsection{Indoors and Outdoors Experiments}
\begin{figure*}
    \centering
    \includegraphics[width=\linewidth]{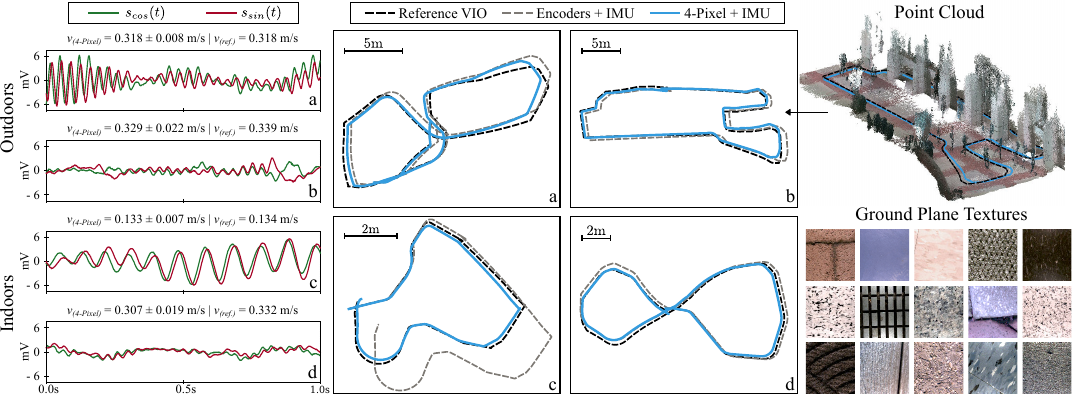}
    \vspace{-0.25in}
    \caption{\textbf{Indoor and outdoor experiments.} On the left, the quadrature signals $s_{cos}(t)$ and $s_{sin}(t)$ of our 4-Pixel sensor over a 1-second window, sampled from the corresponding trajectories (in the center), for strong (a, c) and weak (b, d) signal amplitudes $A(t)$. The speed and uncertainty predicted by the TCN are reported above each panel, together with the reference speed. The trajectories (solid blue) computed using our minimalist odometry sensor paired with an IMU closely follow the reference VIO trajectories (dashed black) across a variety of outdoor (a-b) and indoor (c-d) environments. They outperform the wheel encoder baseline (dashed gray) in all cases. The point cloud corresponding to path `b' is shown on the top-right. Some of the arbitrary real-world textures (brick, marble, concrete, carpet, pebbles, wood planks, machined metal), captured by an auxiliary camera mounted alongside our sensor are shown on the bottom-right.}\label{fig:results}
    \vspace{-0.2in}
\end{figure*}
We have evaluated the performance of our minimalist odometry sensor by teleoperating the robot in diverse indoor and outdoor environments, totaling 920~meters of travel distance and 87~minutes of travel time over 11 indoor trajectories~(618~m in 61~min) and 5 outdoor trajectories~(306~m in 26~min). During teleoperation, we only controlled the direction of the robot while its speed was randomly varied. 

We processed the $1~\text{kHz}$ sensor signals offline to evaluate various update rates of the TCN predictions.
Specifically, we applied a 1-second sliding window of 1000 samples, using strides of 1, 10, 33, and 67 milliseconds to update the speed estimates at $1~\text{kHz}$, $100~\text{Hz}$, $30~\text{Hz}$, and $15~\text{Hz}$ respectively.
We used the network's uncertainty to drop low-confidence estimates, then a median filter to remove transient outliers.
Finally, we fused this filtered speed with calibrated gyroscope readings from the IMU of an Intel RealSense D455 camera to compute planar odometry.

To obtain the reference trajectories we processed visual, depth, and inertial data from the RealSense camera using RTAB-Map~\cite{labbe2019rtab}, with loop closure and pose-graph optimization. Most of our trajectories are loops, which constrains the pose graph, and the metric scale is fixed by the calibrated stereo baseline of the D455. RTAB-Map on visual-inertial-depth data achieves an ATE of 
$2.9\pm2.0~\text{cm}$ on EuRoC and $7.6\pm4.3~\text{cm}$ on TUM RGB-D, both challenging benchmarks involving more demanding 6-DoF motion than our planar setting. The reference VIO trajectories are thus not an absolute ground truth but a high-fidelity, million-pixel upper bound against which we validate our 4-pixel sensor.

We compared our sensor with wheel encoders, which represent the standard baseline for differential-drive robots.
In particular, we considered two specific configurations. 
The first, \textit{Encoders}, computes planar odometry by using the mean and differential measurements from the robot's four wheel encoders to derive linear and angular speed. The second, \textit{Encoders + IMU}, retains the encoder-based linear speed but, for angular speed, it integrates the same calibrated gyroscope readings used in our approach. This second configuration serves as a direct comparison with our sensor, as they both are being relied on for their linear speed estimates.

\begin{table}[t]
    \centering
    \caption{Minimalist Odometry Performance}
    \vspace{-0.1in}
    \label{tab:real_results}
    \setlength{\tabcolsep}{2pt}
    \begin{tabular}{@{}l c c c c@{}}
        \toprule
        & \multicolumn{2}{c}{\textbf{Indoor}} & \multicolumn{2}{c}{\textbf{Outdoor}} \\
        \cmidrule(lr){2-3} \cmidrule(l){4-5}
        \textbf{Method} & \textbf{ATE\,(m)$\downarrow$} & \textbf{Drift\,(\%)$\downarrow$} & \textbf{ATE\,(m)$\downarrow$} & \textbf{Drift\,(\%)$\downarrow$} \\
        \midrule
        Encoders& 9.75\,{\scriptsize$\pm$8.45} & 25.60\,{\scriptsize$\pm$30.00} & 14.09\,{\scriptsize$\pm$7.74} & 30.25\,{\scriptsize$\pm$26.13} \\
        Encoders\,+\,IMU& 0.75\,{\scriptsize$\pm$0.47} & 1.62\,{\scriptsize$\pm$1.66} & 0.74\,{\scriptsize$\pm$0.19} & 1.37\,{\scriptsize$\pm$1.20} \\
        \midrule
        4-Pixel\,+\,IMU\,(1\,kHz)& \textbf{0.28\,{\scriptsize$\pm$0.11}} & \textbf{0.60\,{\scriptsize$\pm$0.22}} & \textbf{0.42\,{\scriptsize$\pm$0.08}} & \textbf{0.62\,{\scriptsize$\pm$0.25}} \\
        4-Pixel\,+\,IMU\,(100\,Hz)& 0.29\,{\scriptsize$\pm$0.12} & 0.65\,{\scriptsize$\pm$0.24} & 0.43\,{\scriptsize$\pm$0.07} & 0.72\,{\scriptsize$\pm$0.12} \\
        4-Pixel\,+\,IMU\,(30\,Hz)& 0.28\,{\scriptsize$\pm$0.11} & 0.65\,{\scriptsize$\pm$0.21} & 0.44\,{\scriptsize$\pm$0.08} & 0.85\,{\scriptsize$\pm$0.22} \\
        4-Pixel\,+\,IMU\,(15\,Hz)& 0.32\,{\scriptsize$\pm$0.13} & 0.67\,{\scriptsize$\pm$0.22} & 0.45\,{\scriptsize$\pm$0.08} & 1.08\,{\scriptsize$\pm$0.28} \\
        \bottomrule
    \end{tabular}
    \vspace{-0.05in}
\end{table}

To quantify the performance of our system, we consider the absolute trajectory error~(ATE) and the percentage of endpoint drift~\cite{carlone2025slam}. 
The ATE represents the global geometric consistency of the path by computing the root-mean-square difference between the position estimates and the reference trajectories. The endpoint drift measures the accumulated relative error over the total traveled distance. 

The quantitative results, summarized in Table~\ref{tab:real_results}, show that our 4-Pixel approach outperforms the baselines for both the indoor and outdoor trajectories.
As expected, the performance of our approach degrades slightly outdoors, as uneven terrains and drastic lighting variations corrupt the sensor signals.
Reducing the TCN update rate (which corresponds to lower computational load) from $1~\text{kHz}$ to $100~\text{Hz}$ and $30~\text{Hz}$ causes a negligible drop in odometry performance, which becomes more pronounced at $15~\text{Hz}$.

The performance of our system depends on the SNR of the measured differential signals. This SNR is determined by the light level in the environment and the visual contrast of the texture. During the 920 m travel, the signal amplitude falls below the read noise $\sigma_r$ for 330 m and below $\sigma_r / 2$ for 150 m. Within these intervals, the speed MAE degrades from $0.026~\nicefrac{\text{m}}{\text{s}}$ when the amplitude is above $\sigma_r$ to $0.042~\nicefrac{\text{m}}{\text{s}}$ below $\sigma_r$ and $0.050~\nicefrac{\text{m}}{\text{s}}$ below $\sigma_r / 2$. Note, however, that the TCN outputs a confidence for each speed prediction, allowing us to discard low-confidence estimates before computing odometry. This is possible, in part, since the SNR can be measured using the differential signal's amplitude.

Fig.~\ref{fig:results} shows the sensor signals and trajectories computed in four of our indoor and outdoor experiments, with the predicted confidence decreasing for the low-amplitude signals. Our system (\mbox{4-Pixel} + IMU) closely tracks the reference VIO for a wide variety of environments with diverse ground textures. Despite the fact that the sensor masks and the TCN were optimized entirely using simulated data, our sensor is able to generalize to arbitrary real-world environments.

\subsection{Comparison with Optical Flow}\label{subsec:optflow}

\begin{table}[t]
    \centering
    
    \caption{Comparison with Camera-Based Optical Flow}
    \vspace{-0.1in}
    \label{tab:optflow}
    \setlength{\tabcolsep}{3pt}
    \begin{tabular}{@{}l c c c@{}}
        \toprule
        \textbf{Resolution} & \textbf{Meas.\,/\,s}$\downarrow$ & \textbf{ATE\,(m)}$\downarrow$ & \textbf{Drift\,(\%)}$\downarrow$ \\
        \midrule
        $256\times256$ & \num{1966080} & 0.23\,{\scriptsize$\pm$0.13} & 0.56\,{\scriptsize$\pm$0.39} \\
        $128\times128$ & \num{491520} & 0.23\,{\scriptsize$\pm$0.12} & 0.54\,{\scriptsize$\pm$0.29} \\
        $64\times64$ & \num{122880} & 0.25\,{\scriptsize$\pm$0.16} & 0.66\,{\scriptsize$\pm$0.32} \\
        $32\times32$ & \num{30720} & 0.62\,{\scriptsize$\pm$0.81} & 1.10\,{\scriptsize$\pm$0.50} \\
        $16\times16$ & \num{7680} & 1.43\,{\scriptsize$\pm$0.83} & 4.67\,{\scriptsize$\pm$3.81} \\
        $8\times8$ & \num{1920} & 2.10\,{\scriptsize$\pm$1.65} & 4.69\,{\scriptsize$\pm$2.78} \\
        $4\times4$ & \num{480} & 8.27\,{\scriptsize$\pm$3.90} & 10.21\,{\scriptsize$\pm$5.01} \\
        $2\times2$ & \num{120} & 8.60\,{\scriptsize$\pm$4.80} & 4.89\,{\scriptsize$\pm$2.45} \\
        \midrule
        \textbf{4-Pixel\,+\,IMU (ours)} & \num{4000} & 0.24\,{\scriptsize$\pm$0.07} & 0.58\,{\scriptsize$\pm$0.15} \\
        \bottomrule
    \end{tabular}
    \vspace{-0.25in}
\end{table}

We compared the performance of our system with an optical flow-based method using $30$ FPS video from an auxiliary downward-facing camera mounted alongside our sensor at the same height and with the same
$70^\circ$ field of view. Using the captured video from six indoor trajectories totaling 309.8 m and 29.4 min, we computed displacement in pixels with a block-matching approach~\cite{honegger2013open} similar to that of commercial optical flow sensors. Then, we converted the displacement to meters using the calibrated camera height and integrated it at $30\,\text{Hz}$ with the same gyroscope measurements used by our system, matching our $30\,\text{Hz}$ configuration in Tab.~\ref{tab:real_results}. We repeated this using videos with progressively lower resolutions and report the results in Tab.~\ref{tab:optflow}.

The performance of our minimalist sensor is comparable to the optical flow-based approach using a $64\times64\,\text{px}^2$ image. This translates to $1024\times$ fewer pixels and $30\times$ fewer measurements per second. Our sensor also remains comparable to full-resolution ($256\times256\,\text{px}^2$) optical flow, which requires $490\times$ more measurements per second, while at lower resolutions the performance of the optical flow-based approach degrades significantly.
\section{Conclusions}\label{sec:conclusions}
We presented a minimalist approach to planar odometry using a sensor comprising just four masked detectors. Theoretical analysis showed Gabor masks were ideal for speed estimation. Through a custom simulator, we jointly optimized the parameters of the Gabor masks and a TCN network. We installed the optimized masks on a prototype sensor and used it with an IMU on a differential drive robot to compute planar odometry. Our minimalist approach achieves robust performance across diverse indoor and outdoor environments.

This work can be extended in several directions. First, we plan to further reduce the cost of our TCN decoder to enable ultra-low-power planar odometry on custom embedded hardware. Second, we would like to explore the use of masks for odometry with more degrees of freedom. This will make our approach applicable to other platforms, such as drones. Finally, minimalist sensing can be exploited to solve other robotics tasks such as terrain classification.

\section{Acknowledgments}\label{sec:conclusions}
This work was supported in part by the Office of Naval Research award N00014-23-1-2096 and in part by the National Science Foundation (NSF) and Center for Smart Streetscapes (CS3) under NSF Cooperative Agreement No. EEC-2133516. Jeremy Klotz was supported by a National Defense Science and Engineering Graduate (NDSEG) Fellowship. The authors thank Matthew Beveridge for technical feedback and Dhruv Yalamanchi for helping with experiments.

\bibliographystyle{IEEEtran}
\bibliography{IEEEabrv, bibliography}

\end{document}